\documentclass[journal]{IEEEtran}
\PassOptionsToPackage{table}{xcolor}
\usepackage{xcolor}
\usepackage{amsmath,amsfonts}
\usepackage{array}
\usepackage[caption=false,font=normalsize,labelfont=sf,textfont=sf]{subfig}
\usepackage{textcomp}
\usepackage{stfloats}
\usepackage{url}
\usepackage{verbatim}
\usepackage{graphicx}
\usepackage{hyperref}

\usepackage{algorithm}
\usepackage{algpseudocode}
\usepackage{multirow}
\usepackage{makecell}
\usepackage{arydshln}
\usepackage{float}

\usepackage{siunitx}  
\usepackage{booktabs}  
\usepackage{bm}

\def\BibTeX{{\rm B\kern-.05em{\sc i\kern-.025em b}\kern-.08em
    T\kern-.1667em\lower.7ex\hbox{E}\kern-.125emX}}
\usepackage{balance}

\begin{document}
\title{DeblurSplat: Traditional SfM-free 3D Gaussian Splatting with Event Camera
for Robust Deblurring}
\author{Pengteng Li, Pinhao Song, Weiyu Guo, Huizai Yao, Yunfan Lu$^*$, F. Richard Yu,~\IEEEmembership{Fellow,~IEEE}, and Hui Xiong$^*$,~\IEEEmembership{Fellow,~IEEE}

\thanks{This work was supported in part by the National Key R\&D Program of China (Grant No.2023YFF0725001), in part by the National Natural Science Foundation of China (Grant No.92370204), in part by the guangdong Basic and Applied Basic Research Foundation (Grant No.2023B1515120057), in part by the Key-Area Special Project of Guangdong Provincial Ordinary Universities (2024ZDZX1007).(\textit{$^*$Corresponding authors: Yunfan Lu, Hui Xiong.})}
\thanks{Pengteng Li, Yunfan Lu, Weiyu Guo, Huizai Yao and Hui Xiong are with the Thust of the Artificial Intelligence, The Hong Kong
University of Science and Technology (Guangzhou), Guangzhou 511453, China (email: \{pli807, ylu066, wguo395, hyao032\}@connect.hkust-gz.edu.cn), {xionghui@ust.hk})}. 
\thanks{Pinhao Song is with the KU Leuven (email: \url{pinhao.song@kuleuven.be}).}
\thanks{F. Richard Yu is with Carleton University (e-mail: \url{richard.yu@carleton.ca}).}
}


\maketitle

\begin{abstract}
In this paper, we propose the first traditional Structure-from-Motion (SfM)-free deblurring 3D Gaussian Splatting method via event camera, dubbed \textbf{\textit{DeblurSplat}}.
We address the motion-deblurring problem in two ways. First, we leverage the pretrained capability of the dense stereo module (DUSt3R) to directly obtain accurate initial point clouds from blurred images. Without calculating camera poses as an intermediate result, we avoid the transfer of cumulative errors from inaccurate camera poses to the positions of initial point clouds.
Second, we introduce the event stream into the deblur pipeline for its high sensitivity to dynamic change. By decoding the latent sharp images from the event stream and blurred images, we can provide a fine-grained supervision signal for scene reconstruction optimization. Extensive experiments across a range of scenes demonstrate that DeblurSplat not only excels in generating high-fidelity novel views but also achieves significant rendering efficiency compared to the SOTAs in deblur 3D-GS.

\end{abstract}    
\section{Introduction}
\label{sec:intro}

Scene reconstruction and novel-view synthesis (NVS) from 2D image collections remain fundamental yet challenging problems in computer vision and computer graphics. While point-based representations, such as 3D-GS \cite{3DGS} and NeRF \cite{NeRF}, have achieved impressive results in generating high-fidelity novel views, they typically assume high-quality inputs captured in ideal conditions. Unfortunately, real-world captures often contain motion blur \cite{Bad-gaussians} due to low-light or long-exposure settings, leading to inaccurate geometry estimation across multiple views. As a result, the performance of both NeRF and 3D-GS deteriorates significantly in blurred scenarios.

\begin{figure}[t]
    \centering
    \includegraphics[width=0.95\linewidth]{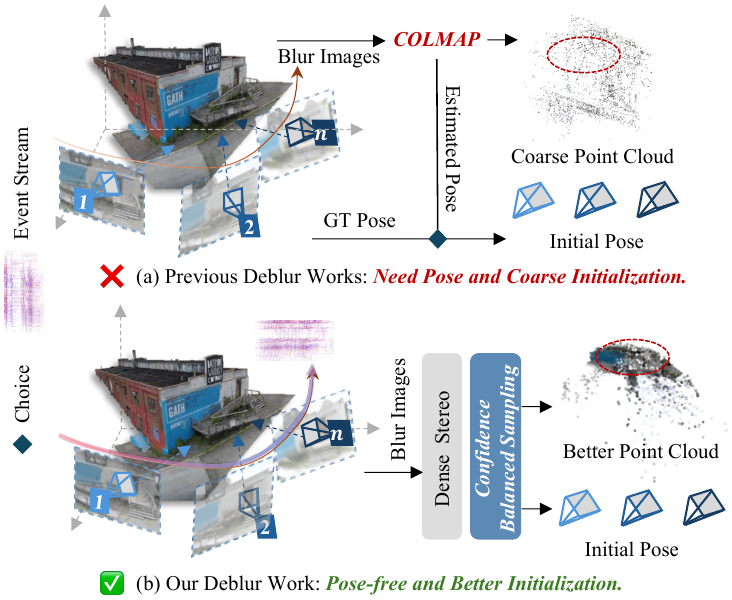}
    \vspace{-0.2cm}
    \caption{Illustration of differences between our work and previous works for deblur reconstruction. (a) Most previous works directly estimate camera poses and point clouds by traditional SfM pipelines like COLMAP \cite{colmap}, leading to coarse point clouds initialization and poses due to the blur geometry uncertainty. (b) Our work mainly utilize sufficient semantic prior from the dense stereo module (DUSt3R \cite{Dust3r}) to initialize better point clouds and camera poses, while leverages event streams to guarantee the overall optimization.}
    \label{fig:abstract}
    \vspace{-0.2cm}
\end{figure}

To address this issue, researchers have explored deblur-based NeRF and 3D-GS approaches that explicitly model the latent physical blur process \cite{Deblur-nerf,Dp-nerf,Bad-nerf,Bad-gaussians}. Recently, the imaging characteristics of event cameras—such as no motion blur and high dynamic range—have made them particularly attractive for deblurring tasks \cite{survey, E2gs,E3NeRF,EvaGaussians,EventSplat,BeSplat,Elite-evgs}. By leveraging the physical imaging process \cite{EDI}, event-based methods can recover clearer scene details, outperforming purely RGB-based pipelines.

Even when event cameras are employed, the overall reconstruction quality is often constrained by the inaccurate initialization provided by traditional Structure-from-Motion (SfM) pipelines (e.g., COLMAP). Motion-blurred images tend to smear out edges and flatten texture details, creating extensive regions of semantically similar pixels \cite{Bad-nerf,EvaGaussians}. This confuses the feature detection and matching process, yielding fewer valid correspondences and a higher likelihood of mismatches. As a result, camera pose estimation is inaccurate, leading to noticeable misalignment across multiple views. Since 3D-GS and NeRF frameworks rely heavily on precise multi-view alignment to fuse geometry and appearance, these pose errors propagate into the final reconstruction, manifesting as distorted geometry, incomplete surfaces, and artifacts in the rendered views. Consequently, the initial point clouds generated by COLMAP are often sparse or inaccurately positioned, further undermining the performance of the entire deblurring pipeline. Using SfM to estimate blur scenes results with several images in longer training time consuming and is prone to failure, which we conduct relative experiments as shown in Fig. \ref{fig:time}. 


Recently, learning-based novel-view synthesis approaches that are independent of traditional SfM pre-processing pipelines (e.g., COLMAP) have emerged~\cite{Light3R,Dust3r,Align3r,FreeSplatter,FLARE,wuyang}. 
Unlike classical SfM methods like COLMAP that rely on detect-and-match of high-frequency local features and optimize camera poses via 2D reprojection error minimization, these approaches estimate scene geometry in an end-to-end manner through dense geometric regression, making them inherently more robust to challenging conditions such as motion blur. 
Among them, DUSt3R~\cite{Dust3r} employs a transformer-based dense stereo model pretrained on large-scale datasets to directly regress 3D coordinates with continuous confidence estimates. 
By capturing global structural layouts from coarse visual cues and maintaining geometric consistency through confidence-aware 3D-to-3D alignment rather than bundle adjustment, DUSt3R remains effective even when local textures and gradients are severely degraded by motion blur. 
These properties motivate us to adopt DUSt3R as a robust geometric initialization module for blurred imagery, where traditional SfM pipelines frequently fail.


Building on these insights, we propose \textbf{DeblurSplat}, a novel motion-deblurring framework for 3D Gaussian Splatting that is independent of traditional SfM pre-processing and robust to SfM failures. 
DeblurSplat integrates event-camera priors with DUSt3R’s learning-based geometric initialization (see Fig.~\ref{fig:abstract}). 
To fully exploit the rich yet noisy point clouds predicted by DUSt3R under blur, we introduce a \textit{Confidence Balanced Sampling} strategy that jointly considers prediction confidence and global spatial coverage, preventing both over-concentration in high-confidence regions and excessive pruning of informative low-confidence structures. 
Furthermore, to address residual artifacts introduced by motion blur, we propose \textit{Progressive Alignment}, which leverages the high temporal resolution of event streams~\cite{E2gs,E3NeRF,EvaGaussians,EventSplat,BeSplat,Elite-evgs} to iteratively refine latent sharp appearances and camera motion, enabling coarse-to-fine optimization toward a geometrically consistent reconstruction. To comprehensively evaluate robustness under diverse blur conditions, we construct seven challenging synthetic scenes in Blender with varying textures, lighting conditions, and object configurations, extending beyond the complexity of existing datasets~\cite{Deblur-nerf}. 
We further collect real-world event-based datasets with controlled motion blur by adjusting camera exposure times, enabling systematic evaluation across multiple blur levels.
Finally, our contributions are summarized as follows:

\begin{itemize}
    \item We propose DeblurSplat, a traditional SfM-free 3D-GS framework designed for robust deblurring, which effectively integrates the extensive prior knowledge from DUSt3R and the high dynamic range information provided by event cameras. To the best of our knowledge, this work constitutes the first attempt to leverage a dense stereo module to tackle the challenge of motion-deblurring in 3D reconstruction.
    \item We propose an initial point clouds sampling strategy named Confidence Balanced Sampling that leverages DUSt3R’s confidence estimates while preserving necessary spatial coverage.
    \item We propose Progressive Alignment to combine the cues from event stream for an accurate scene reconstruction. By estimating the latent sharp images and latent camera poses, we provide a fine-grained guidance for scene reconstruction optimization.
    \item Extensive experiments on diverse scenes demonstrate that DeblurSplat achieves SOTA results and outperforms motion-deblur reconstruction counterparts significantly.
\end{itemize}




\section{Related Work}
\label{sec:Related}

\begin{figure*}[t]
    \centering
    \includegraphics[width=0.95\linewidth]{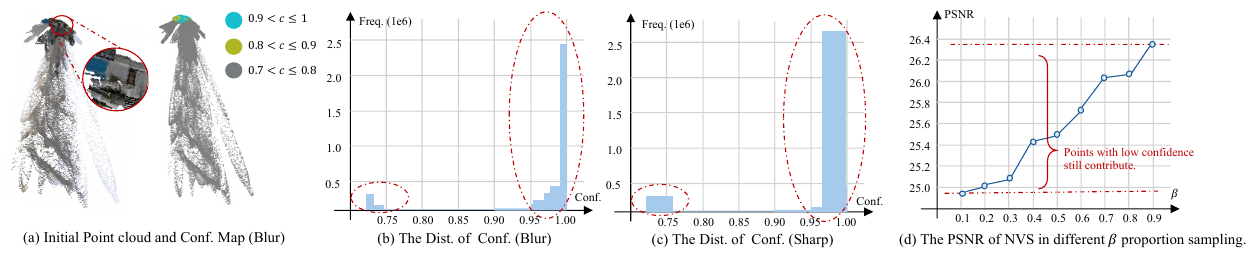}
    \vspace{-0.4cm}
    \caption{Empirical study of initial point clouds generated from blur images by DUSt3R. Here, ``Freq.'' denotes frequency, ``Dist.'' denotes distribution, ``Conf.'' denotes confidence. (d) means that we sample the points that have a confidence level greater than their corresponding quantile threshold of $1-\beta$, which we sample a proportion of the point cloud equal to $\beta$ in total.}
    \label{fig:analysis}
    \vspace{-0.2cm}
\end{figure*}

\subsection{Motion-Deblur Reconstruction}

Motion blur is a frequent occurrence in real-world imaging scenarios and poses a significant challenge for accurate 3D scene reconstruction. Efforts to address this issue have progressed in both the RGB domain and the event-based domain. In the RGB domain, early methods \cite{Deblur-nerf,Dp-nerf} learned a blur formation kernel to approximate the image blurring process, while subsequent approaches \cite{Bad-gaussians,Bad-nerf} integrated blur modeling with bundle adjustment for deblurring. Nevertheless, even when synthesizing motion blur, recovering precise texture details across multiple views remains non-trivial.
By contrast, event cameras provide asynchronous event streams with several notable advantages, including a high dynamic range ($>$120dB vs. 60dB in conventional sensors) \cite{survey}, which has attracted considerable attention for deblurring. Early event-based methods \cite{EventNerf,E-NeRF,Event3DGS,Event-3DGS} convert event streams into intensity images or temporal differences, using these signals to guide novel-view synthesis. 
Later works \cite{Robust-e-NeRF,Deblur-e-NeRF,SweepEvGS} incorporate additional event-imaging factors, such as event-motion. 
More recently, approaches \cite{E2gs,E2nerf,BeSplat,Ev-GS} integrate motion-blur modeling \cite{Bad-gaussians,Bad-nerf} with event-camera decoupling \cite{EvaGaussians,Ev-DeblurNeRF} to jointly optimize latent poses. EventSplat \cite{EventSplat} excels in utilizing high-temporal resolution event streams to guide Gaussian densification, while DiET-GS \cite{DiET-GS} leverages generative diffusion priors for latent frame recovery.

Despite these advances, most existing event-based deblurring and 3DGS reconstruction methods still rely on conventional SfM pipelines for camera pose estimation and initial geometry recovery. However, under severe motion blur, SfM is highly susceptible to failure due to the loss of discriminative textures and edge structures, often leading to inaccurate poses and error accumulation in the reconstructed point clouds. In practice, these methods frequently require the assistance of sharp frames for successful initialization, which limits their autonomy. Moreover, the latent-pose optimization process further introduces substantial computational overhead, as evidenced by prior works such as~\cite{E2gs,E2nerf,EvaGaussians}, where pose estimation constitutes a significant portion of the overall runtime. In contrast, our approach eliminates the dependence on SfM entirely by enabling robust initialization directly from blurred inputs, thereby improving both reliability and efficiency.

\vspace{-0.2cm}
\subsection{SfM-free Novel-view Synthesis}
Recent advances in novel-view synthesis from uncalibrated images have been greatly facilitated by progress in Multi-view Stereo (MVS) methods \cite{LiftImage3D,FLARE,FreeSplatter,Fast3R}. Cutting-edge MVS approaches \cite{Dust3r,MASt3R,FLARE} can directly estimate the geometry of visible surfaces without requiring explicit camera parameters. For instance, PF3plat \cite{Pf3plat} aligns 3D Gaussians coarsely by leveraging pre-trained models for monocular depth estimation and visual correspondence. Meanwhile, Splatt3R \cite{Splatt3r} and NopoSplat \cite{NoPoSplat} employ DUSt3R \cite{Dust3r} and MASt3R \cite{MASt3R} to predict point maps as proxy geometry, subsequently learning 3D Gaussians for sparse-view reconstruction. Although these methods primarily explore generalization in sparse-view 3D reconstruction, our work is the first to fully exploit dense stereo techniques for deblurring reconstruction and novel-view synthesis, thus extending their applicability to challenging blur scenarios.

\begin{table}[t]
\centering
\caption{Quantitative analysis of DUSt3R's \cite{Dust3r} geometric degradation across different blur levels. Poses estimated from sharp images ($K=1$) are used as reference (GT) for computing relative ATE and RPE.}
\resizebox{0.75\linewidth}{!}{
\begin{tabular}{l c c c}
\toprule
Blur Level & $K$ & ATE (cm)$\downarrow$ & RPE (cm)$\downarrow$ \\
\midrule
Sharp          & 1   & --- (Ref.) & --- (Ref.) \\
Mild Blur      & 15  & 1.35 & 0.72 \\
Moderate Blur  & 30  & 2.12 & 1.18 \\
Severe Blur    & 51  & 2.95 & 1.62 \\
Extreme Blur   & 75  & 3.42 & 1.95 \\
\bottomrule
\end{tabular}}
\vspace{-0.2cm}
\label{tab:blur_robustness_analysis}
\end{table}
\section{Method}

\begin{figure*}[t]
    \centering
    \includegraphics[width=0.85\linewidth]{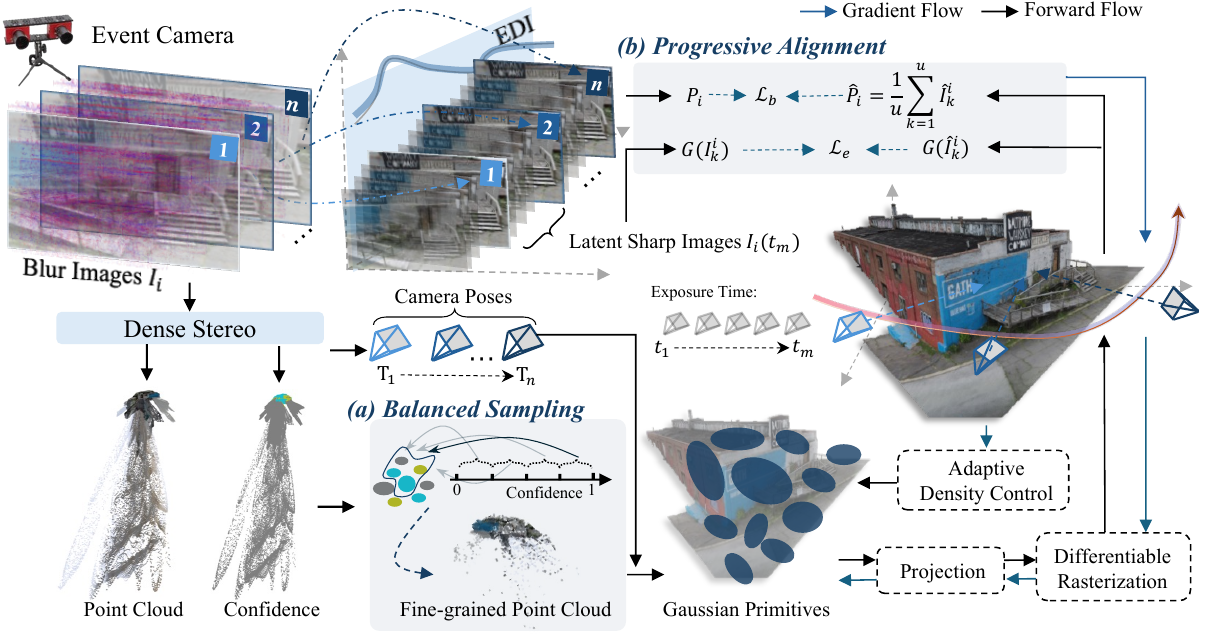}
    \vspace{-0.2cm}
    \caption{Illustration of our proposed DeblurSplat. We first input the blur images into the dense stereo module (DUSt3R \cite{Dust3r}) for getting estimated point clouds and corresponding confidence. Then, we propose the Confidence Balanced Sampling to extract high-quality point clouds by conforming the original points distribution. In the mean time, we utilize the event stream to decouple blur images into multiple latent sharp images by EDI \cite{EDI}. Finally, we perform Progressive Alignment to recover the clear semantics in coarse to fine manner.}
    \vspace{-0.2cm}
    \label{fig:framework}
\end{figure*}

The proposed DeblurSplat pipeline is illustrated in Fig. \ref{fig:framework}. The following two subsections detail its main components. In Subsection \ref{subsec:cbs}, we generate dense point clouds with associated confidence values via DUSt3R, then analyze various sampling strategies to address potential issues. Inspired by the analysis, we introduce Confidence Balanced Sampling, which employs confidence scores to exclude unreliable points while preserving essential spatial coverage. In Subsection \ref{subsec:pa}, we utilize event streams to recover both latent sharp images and latent motion poses, providing fine-grained optimization cues for scene reconstruction.

\vspace{-0.2cm}
\subsection{Sampling Initial Point Clouds from DUSt3R}
\label{subsec:cbs}

\noindent \textbf{Motivation}. As outlined in Section \ref{sec:intro}, DUSt3R generates initial point clouds with associated confidence scores by converting multi-view images into a 3D representation. In contrast to COLMAP, which requires camera pose estimation directly from blurred images, DUSt3R can alleviate pose-related inaccuracies, thus producing more reliable point clouds. However, existing SfM-free novel-view synthesis approaches \cite{SPARS3R,NoPoSplat,Splatt3r} generally operate on sharp images, prompting two crucial questions for blurred-image scenarios: \textit{(1) To what extent can we rely on DUSt3R’s point clouds? (2) How can we effectively utilize these point clouds for deblur reconstruction?}

Regarding the first problem about the robustness of DUSt3R as a coarse initializer under motion blur, we conduct a controlled study by progressively increasing the blur severity. As shown in Table~\ref{tab:blur_robustness_analysis}, poses estimated from sharp images ($K=1$) are treated as a pseudo ground truth to measure the relative trajectory drift induced by blur. Although increasing $K$ degrades fine-scale accuracy, the estimated poses remain within a stable convergence basin, which is sufficient for subsequent joint optimization.
This behavior can be explained by the nature of motion blur as a temporal integration process. Let the latent sharp image at time $t$ be $I(x,t)=I_0(x-u(t))$, where $u(t)$ denotes camera or scene motion. Averaging $K$ latent frames yields
\begin{equation}
    \bar{I}(x)\approx\frac{1}{T}\int_{0}^{T} I_0(x-u(t))\,dt,
\end{equation}
which acts as a motion-dependent low-pass filter that suppresses high-frequency details while preserving mid-to-low frequency geometric statistics, such as global layout and relative depth ordering. Accordingly, we employ DUSt3R’s Global Aligner to obtain a globally consistent but coarse point cloud initialization by optimizing relative transformations across all view pairs. While motion blur weakens local texture correspondences, DUSt3R’s transformer-based architecture captures motion-invariant structural cues through long-range self-attention, providing sufficient constraints to stabilize the global scale. We emphasize that this output is not assumed to be accurate, but serves as a noisy yet structurally coherent prior for subsequent optimization.

To address the second question regarding how to effectively leverage the coarse point cloud, a naive strategy is to directly utilize all points predicted by DUSt3R. However, as illustrated in Fig.~\ref{fig:analysis}(a, b), a large portion of high-confidence points concentrates within limited overlapping regions of the input views. This over-representation leads to inefficient Gaussian primitive initialization and increased computational cost~\cite{Instantsplat,Pointnet++}. Moreover, erroneous points induced by motion blur are indiscriminately included, which degrades novel-view synthesis accuracy. An alternative strategy is to retain only high-confidence points, under the assumption that they are more reliable in blurred scenarios. Nevertheless, our empirical analysis (Fig.~\ref{fig:analysis}(b--d)) reveals that this approach is also suboptimal. First, the confidence histograms of sharp and blurred inputs exhibit similar distributions, with substantial mass around moderate confidence values (e.g., $\sim$0.75), indicating that confidence alone is insufficient to distinguish erroneous points. Second, when reconstructing scenes with BAD-GS~\cite{Bad-gaussians}, we observe that relaxing the confidence threshold and incorporating additional low-confidence points can actually improve reconstruction quality. This suggests that low-confidence points still encode essential spatial structure, and aggressively discarding them results in a loss of geometric completeness. Motivated by these observations, we propose \textit{Confidence Balanced Sampling}, a principled strategy that leverages DUSt3R’s confidence estimates while explicitly preserving spatial coverage. By balancing the contribution of points across confidence levels, our method mitigates blur-induced noise without sacrificing the structural integrity required for accurate deblurring and novel-view synthesis.

\noindent \textbf{Confidence Balanced Sampling}.
Given blurry image sets $ P_i \in \mathbb{R}^{H \times W \times 3} $ where $ i \in \{1,2,\ldots,N\} $, we process these images through DUSt3R to generate initial coarse point clouds $ \mathcal{P} \in \mathbb{R}^{K \times 3} $ and their associated confidence scores $ \mathcal{C} \in \mathbb{R}^K $, where $ K $ denotes the number of points. The goal of Confidence Balanced Sampling is to sample $L$ points as Gaussian primitives for scene reconstruction.
We partition the confidence range $ [c_{\text{min}}, c_{\text{max}}] $ into $ M $ intervals $ \{B_m\}_{m=1}^M $, defined as $ B_m = [c_{\text{min}} + (m-1)\Delta, c_{\text{min}} + m\Delta) $ with $ \Delta = \frac{c_{\text{max}} - c_{\text{min}}}{M} $. For each interval $ B_m $, we extract the subset $ \mathcal{S}_m = \{ \mathbf{p}_k \in \mathcal{P} \,|\, c_k \in B_m \} $, where $ c_k $ is the confidence of point $ \mathbf{p}_k $. The sampling process is designed as:  
\begin{equation}
    \mathcal{P}_{\text{sampled}} = \bigcup_{m=1}^M \left\{ \mathbf{p}_k \,|\, \mathbf{p}_k \sim \mathbb{P}(\mathcal{S}_m), \, \mathbb{P}(\mathbf{p}_k) \propto c_k \right\},
\end{equation}
where points are randomly selected within each $ \mathcal{S}_m $ with probabilities proportional to their confidence. To ensure balanced coverage across confidence levels, we allocate a fixed number of points $ \frac{L}{M} $ per interval $ B_m $. This enforces sparsity in low-confidence regions (prone to noise) and dense sampling in high-confidence regions (critical structures). The resulting subset $ \mathcal{P}_{\text{sampled}} $ provides high-quality initialization for Gaussian primitives in deblurring NVS.

One might consider purely spatial sampling (e.g., Farthest Point Sampling, FPS \cite{Pointnet++,Instantsplat}) to maintain geometric coverage. Although spatial sampling can ensure uniform coverage in clear point clouds, in blurred point clouds it risks selecting outlier clusters (e.g., duplicated or drifted points), further distorting the reconstruction. By contrast, our approach treats DUSt3R’s confidence map as a relative metric in blurred conditions, striking a balance between high-quality point extraction and overall scene fidelity. We provide a comprehensive evaluation of this design in our ablation study (Table \ref{tab:samplingMet}).

\begin{table*}[t]
\centering
\caption{Quantitative comparisons of novel view synthesis across simple, medium, difficult level and real-world scenes. The table reports the average performance for each kind, demonstrating that our method consistently surpasses previous state-of-the-art approaches across all metrics. Best results in \textbf{bold}, second-best \underline{underlined.}}
\vspace{-0.2cm}
\resizebox{0.9\linewidth}{!}{
\begin{tabular}{c|c|cccccccc|c}
\toprule
Scene Type & Metric & B-NeRF & B-3DGS & UFP-GS & EFN-GS & E\textsuperscript{2}GS & BAD-NeRF & BAD-GS & EDNeRF & Ours \\
\midrule
\multirow{3}{*}{Hard} 
& PSNR$\uparrow$ & 21.46 & 22.01 & 23.53 & 23.44 & 23.57 & 24.07 & 23.34 & \underline{24.20} & \textbf{26.94} \\
& SSIM$\uparrow$ & .6222 & .6456 & .6979 & .6998 & .7066 & .7253 & .7408 & \underline{.7657} & \textbf{.8235} \\
& LPIPS$\downarrow$ & .3701 & .3373 & .2775 & .2825 & .2761 & .2325 & .2120 & \underline{.1979} & \textbf{.1355} \\
\midrule

\multirow{3}{*}{Normal} 
& PSNR$\uparrow$ & 22.31 & 22.89 & 26.88 & 26.26  & 28.07 & 29.35 & 31.25 & \underline{31.49} & \textbf{32.09} \\
& SSIM$\uparrow$ & .7467 & .7623 & .8353 & .8155 & .8821 & .9033 & .9255 & \underline{.9389} & \textbf{.9453} \\
& LPIPS$\downarrow$ & .3177 & .2148 & .2084 & .2326 & .1337 & .0998 & .0540 & \underline{.0541} & \textbf{.0406} \\
\midrule

\multirow{3}{*}{Real} 
& PSNR$\uparrow$ & 15.18 & 16.69 & 18.98 & 19.57 & 20.61 & 21.47 & 22.42 & \underline{22.57} & \textbf{25.93} \\
& SSIM$\uparrow$ & .5777 & .6200 & .6868 & .6972 & .6863 & .6937 & \underline{.7129} &  .7049 & \textbf{.7995} \\
& LPIPS$\downarrow$ & .3870 & .3296 & .2289 & .2135 & .2124 & .2084 & .1972 &  \underline{.1940} & \textbf{.0840} \\
\bottomrule
\end{tabular}}
\vspace{-0.4cm}
\label{tab:all}
\end{table*}

\subsection{Progressive Alignment}
\label{subsec:pa}
\noindent \textbf{Blur Images Decouple.} 
Event-based deblurring methods \cite{EvaGaussians,E2VID,Event-3DGS,Robust-e-NeRF} have recently garnered significant attention. By leveraging the continuous event stream captured alongside standard RGB frames, these approaches can reconstruct high-fidelity scenes even under challenging conditions such as low-light or motion blur. Motivated by these advances, we introduce event streams into our deblurring pipeline as a high-dynamic-range semantic prior for guidance. Specifically, following the strategy of previous works, we adopt the Event-based Double Integral (EDI) \cite{EDI} to decouple motion-blurred images using event data. Detailed event-camera imaging principles are omitted here for brevity and can be found in the supplementary material.

In detail, a motion-blurred image $P_i$ is acquired over the exposure interval $[t_0, t_u]$ in tandem with corresponding event bins $\{E_k\}^u_{k=1}$. We denote the latent sharp image at time $t_1$ is $I_1$. According to the event imaging process, the sharp image $I_k$ at time $t_k$ within each event bin can be expressed as:
\begin{equation}
    I_k = I_0 e^{\Theta\sum^{k}_{i=1}E_i}, k>0,
\end{equation}
where $\Theta$ is the event-camera response threshold.
By assuming that the blur in $P_i$ is a time-weighted combination of multiple images \cite{survey}, and given that the exposure time is uniformly divided into $u$ segments, $P_i$ can be approximated as the average of these latent images:
\begin{equation}
    P_i = \frac{I_0}{u+1}(1+e^{\Theta\sum^{1}_{i=1}E_i}+\cdots+e^{\Theta\sum^{u}_{i=1}E_i}).
\label{eq:blur}
\end{equation}
From this, we can further infer $I_0$ and $I_k$ at any moment $k$:
\begin{equation}
\begin{aligned}
    I_0 &= \frac{(u+1)P_i}{1+e^{\Theta\sum^{1}_{i=1}E_i} + \cdots + e^{\Theta\sum^{u}_{i=1}E_i}}, \\
    I_k &= \frac{(u+1)P_i e^{\Theta\sum^{k}_{i=1}E_i}}{1+e^{\Theta\sum^{1}_{i=1}E_i} + \cdots + e^{\Theta\sum^{u}_{i=1}E_i}}.
\end{aligned}
\end{equation}
Accordingly, each blurred image $P_i$ can be decomposed into a set of latent sharp images $\{I^i_k\}^u_{k=1}$, each containing richer texture information.

\noindent \textbf{Latent Motion Modeling.} 
During the exposure period, a motion-blurred image can be viewed as a sequence of latent images captured along a camera trajectory, which can be discretized by poses $\{\boldsymbol{T}^i_k\}^u_{k=1} \in \textbf{SE}(3)$. We thus parameterize and optimize $\{\boldsymbol{T}^i_k\}^u_{k=1}$ in a bundle-adjustment framework \cite{IncEventGS} to jointly recover the camera trajectory responsible for blur formation. Concretely, we parameterize $\Delta\boldsymbol{T}^i_1$ and $\Delta\boldsymbol{T}^i_u$ and linearly interpolate intermediate poses in the Lie algebra of $\textbf{SE}(3)$, where $\Delta\boldsymbol{T}^i$ represents the pose adjustment. Hence, at any moment $k$, the actual pose $\boldsymbol{T}^i_k$ for the corresponding latent image $I^i_k$ is: 
\begin{equation}
    \boldsymbol{T}^i_k = \hat{\boldsymbol{T}^i} \cdot \Delta\boldsymbol{T}^i_1 \cdot \text{Exp}(\frac{k}{u} \cdot \text{Log}((\Delta\boldsymbol{T}^i_1)^{-1} \Delta\boldsymbol{T}^i_u)
    \label{eq:linearpose}
\end{equation}
where $k \in \{1,2,...,u\}$ and $\hat{\boldsymbol{T}^i}$ denotes the initial pose for blur images $P_i$. We optimize $\Delta\boldsymbol{T}^i_1$ and $\Delta\boldsymbol{T}^i_u$ jointly with the learnable parameters of the Gaussian primitives. Note that we also refine the original pose for blurred images, as the DUSt3R-generated poses \cite{Instantsplat,SPARS3R} may be suboptimal. Eq. \ref{eq:linearpose} models the camera motion during exposure using linear interpolation in the Lie algebra of $\mathrm{SE}(3)$, which corresponds to a first-order approximation of a continuous trajectory over a short temporal window. Given the typically brief exposure duration of a single blurred frame, higher-order motion components (e.g., acceleration) have limited influence on the integrated image formation, making this approximation sufficient in practice. This modeling choice is widely adopted in motion-deblurring model \cite{Bad-gaussians,Bad-nerf}, as it offers a favorable balance between expressiveness and optimization stability. In our framework, we assume the residual deviation from constant-velocity motion is further compensated by dense event supervision (Eq. \ref{eq:opm}) and the anisotropic flexibility of 3D Gaussian primitives, ensuring robust pose and geometry recovery.

\noindent \textbf{Event Guided Optimization.} In each training iteration, we simultaneously render $u$ latent sharp images $\{\hat{I^i_k}\}^u_{k=1}$ from 3D-GS along the camera trajectories $\boldsymbol{T}^i_k$ corresponding to each blurred view $P_i$. A direct strategy would be to align these renders $\{\hat{I^i_k}\}^u_{k=1}$ with the actual latent images $\{I^i_k\}^u_{k=1}$. However, despite containing more texture details than $P_i$, these renders do not fully recover ideal latent frames and often exhibit low visual fidelity (especially in color), thereby degrading optimization outcomes. To mitigate this, we transform the renders into grayscale for alignment, thereby removing color interference while retaining sufficient texture and shape cues:
\begin{equation}
    \mathcal{L}_e = \frac{1}{u} \sum^u_{k=1} \parallel G(\hat{I^i_k}) - G(I^i_k) \parallel_1,
    \label{eq:fine}
\end{equation}
where $G(\cdot)$ denotes the standard transfer function for gray space:

\begin{equation}
    G(I) = 0.299 \cdot R + 0.587 \cdot G + 0.114 \cdot B, 
\end{equation}
where $R$, $G$, and $B$ denote the red, green, and blue color channels of the input image $I$. The choice of this standard formula is based on its effectiveness in preserving the structural and textural information required for our Event-Guided Optimization, providing sufficient semantic cues while being computationally efficient. Next, to simulate the formation of motion-blurred images, we approximate Eq.~\ref{eq:blur} by averaging the $u$ latent renders: $\hat{P^i}=\frac{1}{u} \sum^u_{k=1} \hat{I^i_k}$ and adopt the same reconstruction loss as in the original 3D-GS \cite{3DGS}:
\begin{equation}
    \mathcal{L}_b = (1-\lambda_b) \cdot \parallel \hat{P^i} - P^i \parallel_1 + \lambda_b \cdot \mathcal{L}_{\text{D-SSIM}}(\hat{P^i},P^i),
    \label{eq:coarse}
\end{equation}
Our total loss combines the above objectives:
\begin{equation}
    \mathcal{L} = \mathcal{L}_b + \lambda_{e} \mathcal{L}_e.
\label{eq:opm}
\end{equation}
Equation~\ref{eq:fine} promotes convergence toward the correct pose by incorporating the EDI-based event prior. Moreover, it helps the model avoid suboptimal solutions stemming from homogeneous pose updates under blurred-image gradients (Eq.~\ref{eq:coarse}). By jointly enforcing geometric, textural, and color fidelity across multiple views, this overall loss formulation pushes the model’s deblurring performance to new heights. More discussion can be referred to our supplementary.

\section{Experiments}

\begin{table}[t]
\centering
\caption{Quantitative comparisons of novel view synthesis across different level blur of real-world scenes. The table reports the average performance for each kind, demonstrating that our method consistently surpasses previous SOTAs.}
\resizebox{0.9\linewidth}{!}{
\begin{tabular}{ccccccc}
\toprule
Level & Metric & B-3DGS & E\textsuperscript{2}GS & BAD-GS & EDNeRF & Ours \\
\midrule
\multirow{3}{*}{1} 
& PSNR$\uparrow$ & 18.25 & 22.88 & \underline{24.92} & 24.75 & \textbf{26.84}\\ 
& SSIM$\uparrow$ & .6656  & \underline{.7502} & .7511 & .7317 & \textbf{.8272}\\ 
& LPIPS$\downarrow$ & .2728 & \underline{.1437} & .1457 & .1727 & \textbf{.0648} \\ 
\midrule

\multirow{3}{*}{2} 
& PSNR$\uparrow$ & 17.74 & 21.32 & \underline{22.22} & 21.48 & \textbf{26.27}\\ 
& SSIM$\uparrow$ & .6385 & \underline{.6905} & .6965 & .6706  & \textbf{.7858} \\ 
& LPIPS$\downarrow$ & .3061 & .2124 & \underline{.2087} & .2219 &  \textbf{.1060} \\ 
\midrule

\multirow{3}{*}{3} 
& PSNR$\uparrow$ & 15.71 & 20.54 & \underline{21.53} & 22.45 & \textbf{25.90}\\ 
& SSIM$\uparrow$ & .6025 & .6629 &  \underline{.6804} & .7080 & \textbf{.7862}\\ 
& LPIPS$\downarrow$ & .3580 & .2205 & \underline{.2097} & .1979 & \textbf{.0875}\\ 

\midrule

\multirow{3}{*}{4} 
& PSNR$\uparrow$ & 15.06 & 17.71 & \underline{21.00} & 21.57 & \textbf{24.72} \\ 
& SSIM$\uparrow$ & .5732 & .6415 & \underline{.6835} & .6968 & \textbf{.7824}\\ 
& LPIPS$\downarrow$ & .3815 & .2728 & \underline{.2247} & .1994 &\textbf{.0862}\\ 
\bottomrule
\end{tabular}}
\vspace{-0.2cm}
\label{tab:blurlevel}
\end{table}

\begin{figure}[t]
    \centering
    \includegraphics[width=0.85\linewidth]{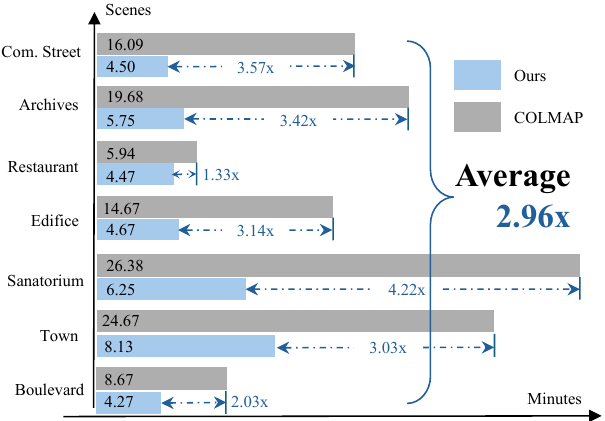}
    \caption{Quantitative comparisons of time consuming (min.) of initialization between COLMAP and DeblurSplat applied in hard scenes. Note that the COLMAP initialization is based on sharp images (almost fail on blur images).}
    \label{fig:time}
    \vspace{-0.2cm}
\end{figure}


\subsection{Implementation Details}

Our model is implemented based on the official code of 3D-GS \cite{3DGS}. Both the optimization of Gaussians and camera poses are performed using the Adam optimizer. The learning rate for Gaussians remains identical to the original 3D-GS. The training process spans 20,000 iterations, with an event reconstruction loss introduced after a 3,000-iteration warm up. We set $m=40$ and $Q_m=125$, which denotes we assume the exposure time of each blur image can be divided into 40 internals and sample a total of 5k points. Then, we set $\lambda_b = 0.2$ and $\lambda_e = 5 \times 10^{-3}$ for balancing the prior guidance. We set $u=10$ and $\Theta=0.27$  following manner of previous works \cite{Bad-nerf,Bad-gaussians,IncEventGS}. All experiments were conducted using a single NVIDIA RTX A6000 GPU. 

For baseline evaluations, we follow the the
standard convention of existing deblurring works (e.g., BAD-GS \cite{Bad-gaussians}, BAD-NeRF \cite{Bad-nerf}), where camera poses are typically initialized by running COLMAP on ground-truth sharp images, which provide them with the best possible ``ideal'' starting point.  Notably, COLMAP consistently fails under blurry conditions as s shown in Fig. \ref{fig:time}. Even when processing sharp images, COLMAP remains computationally expensive whereas our method achieves almost 2.96× faster performance.

\begin{figure*}
    \centering
\includegraphics[width=\linewidth]{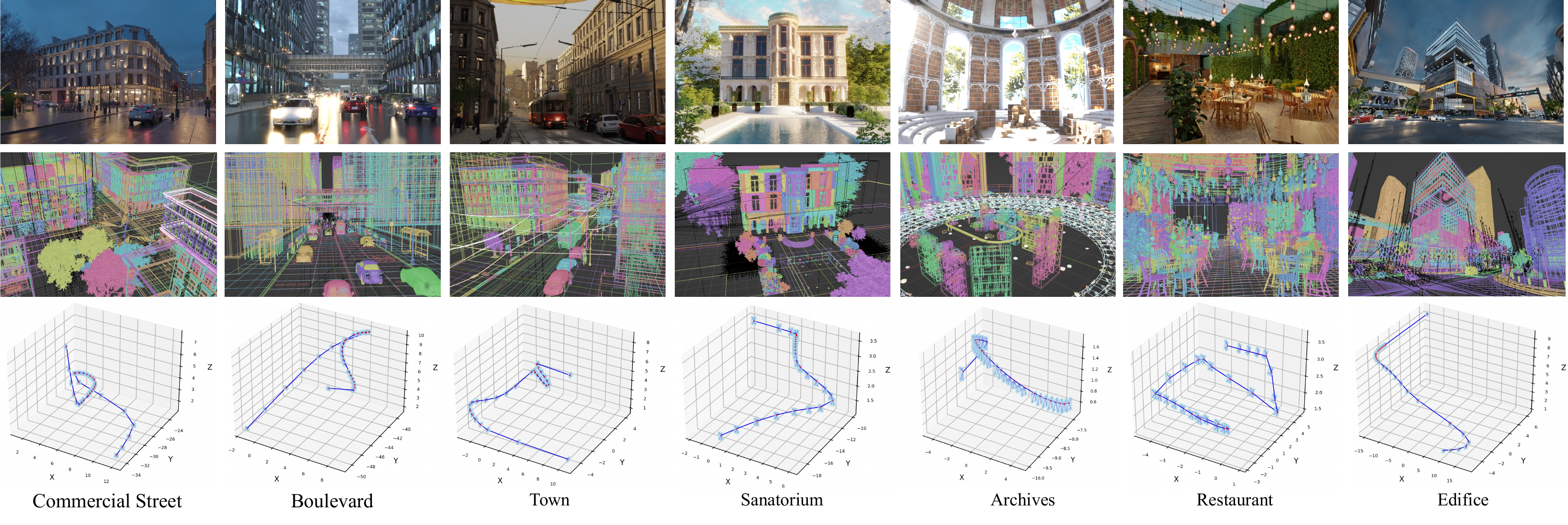}
    \vspace{-0.2cm}
    \caption{Illustration of our crafted hard scene datasets. For each scene, the first row is the rendered mode and the second row is the wire frame mode with random colors. Each scene has diverse and multiple objects and complex lighting involving daylight and night, which is more complicated than synthesis dataset from BAD-NeRF \cite{Bad-nerf}.
    The last row is the corresponding camera trajectory. Compared to normal scene datasets \cite{Deblur-nerf}. The average movement distance of the camera trajectory we proposed is over 10 meters, and the focus view is constantly switched every frame, which poses a great challenge to general deblurring reconstruction models.}
    \label{fig:scene}
\end{figure*}

\begin{figure}
    \centering
    \includegraphics[width=0.7\linewidth]{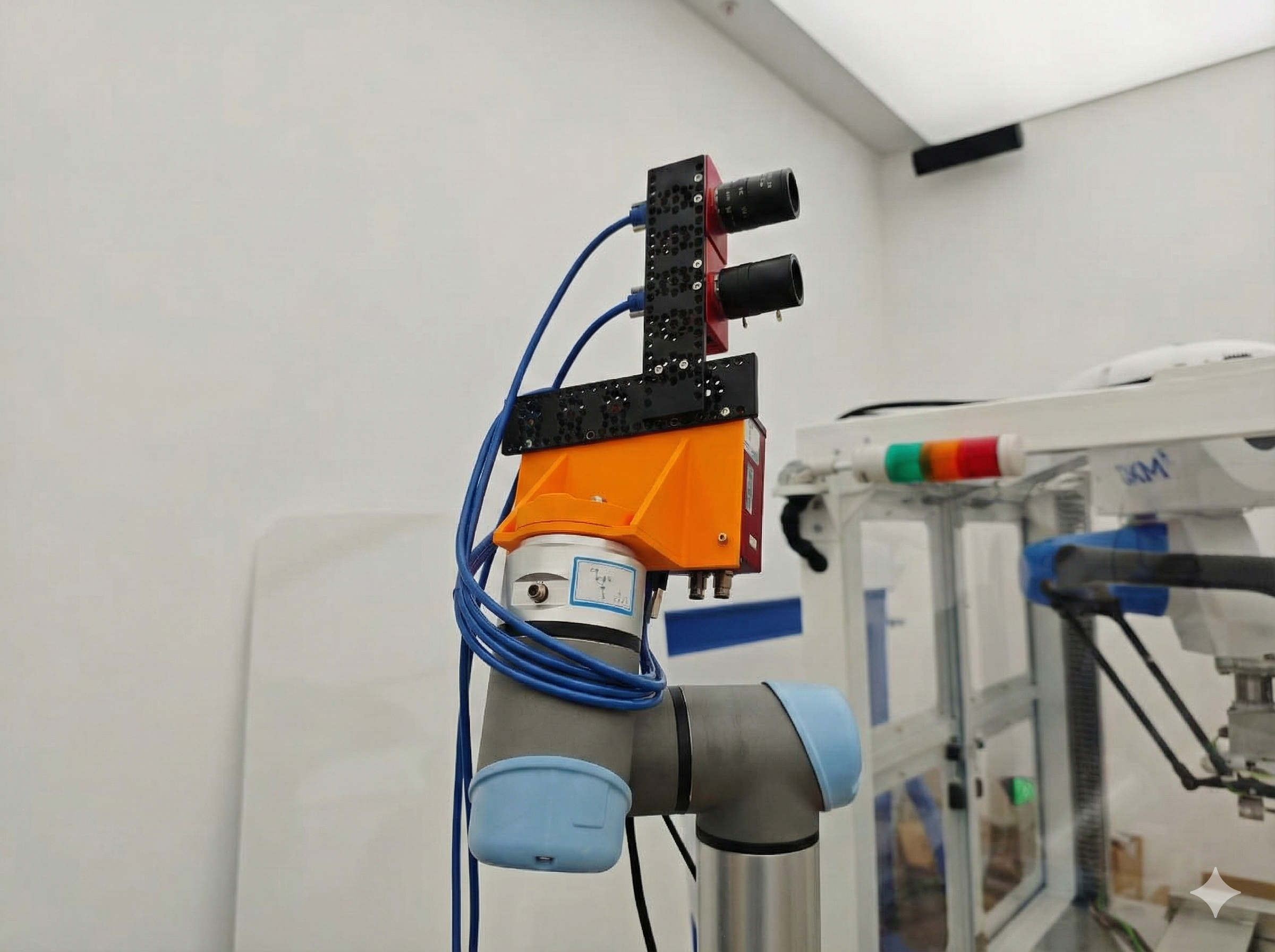}
    \caption{Illustration of our equipment for creating real world datasets. We equip the DAVIS346 event camera on the UR-5E robotic arm.}
    \label{fig:camera}
\end{figure}

\begin{figure*}[t]
    \centering
    \includegraphics[width=0.8\linewidth]{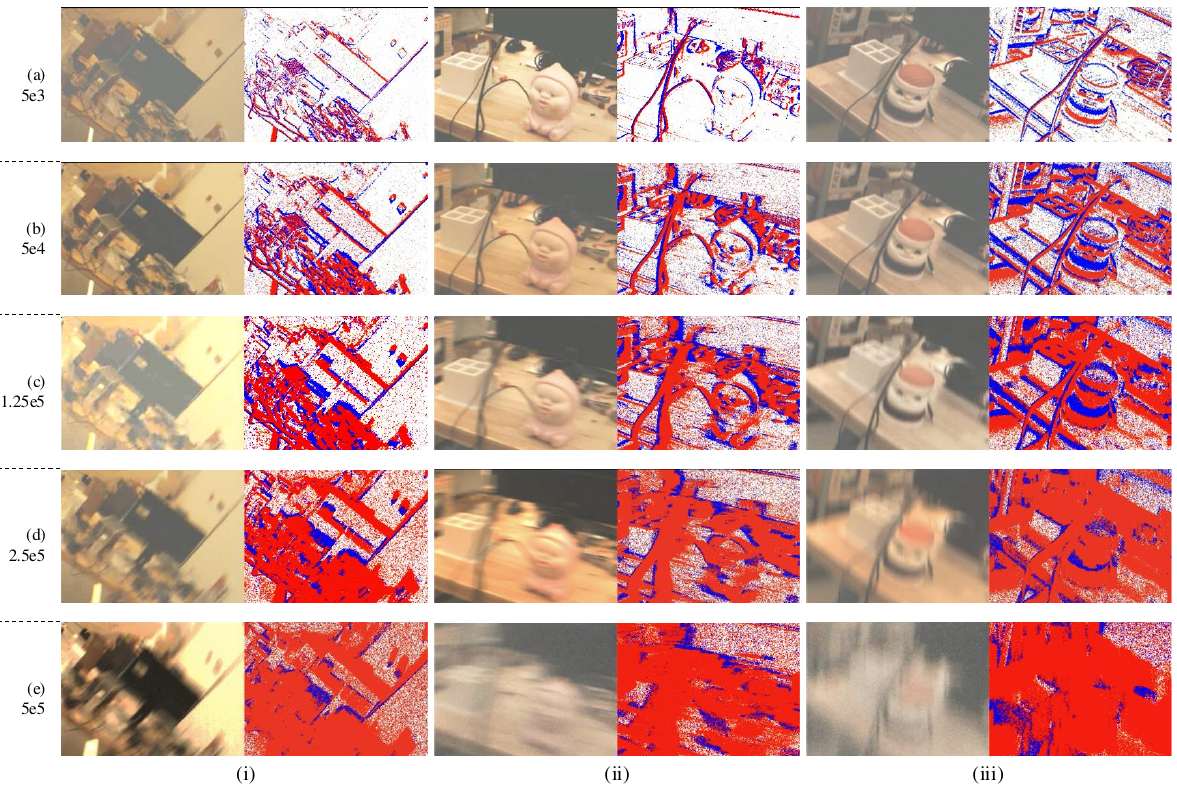}
    \caption{
    Different blur degrees for each scene in the real-world dataset captured using the DVS346 camera. To capture videos with varying blur levels, we controlled the exposure times across different scenes. The dataset contains frames with varying exposure durations, displayed in the following order: (a) 5e3, (b) 5e4, (c) 1.25e5, (d) 2.5e5, and (e) 5e5 $\mu s$. \label{10-RealBlurDVS}
    }
\end{figure*}

\begin{figure*}[t]
    \centering
    \includegraphics[width=0.98\linewidth]{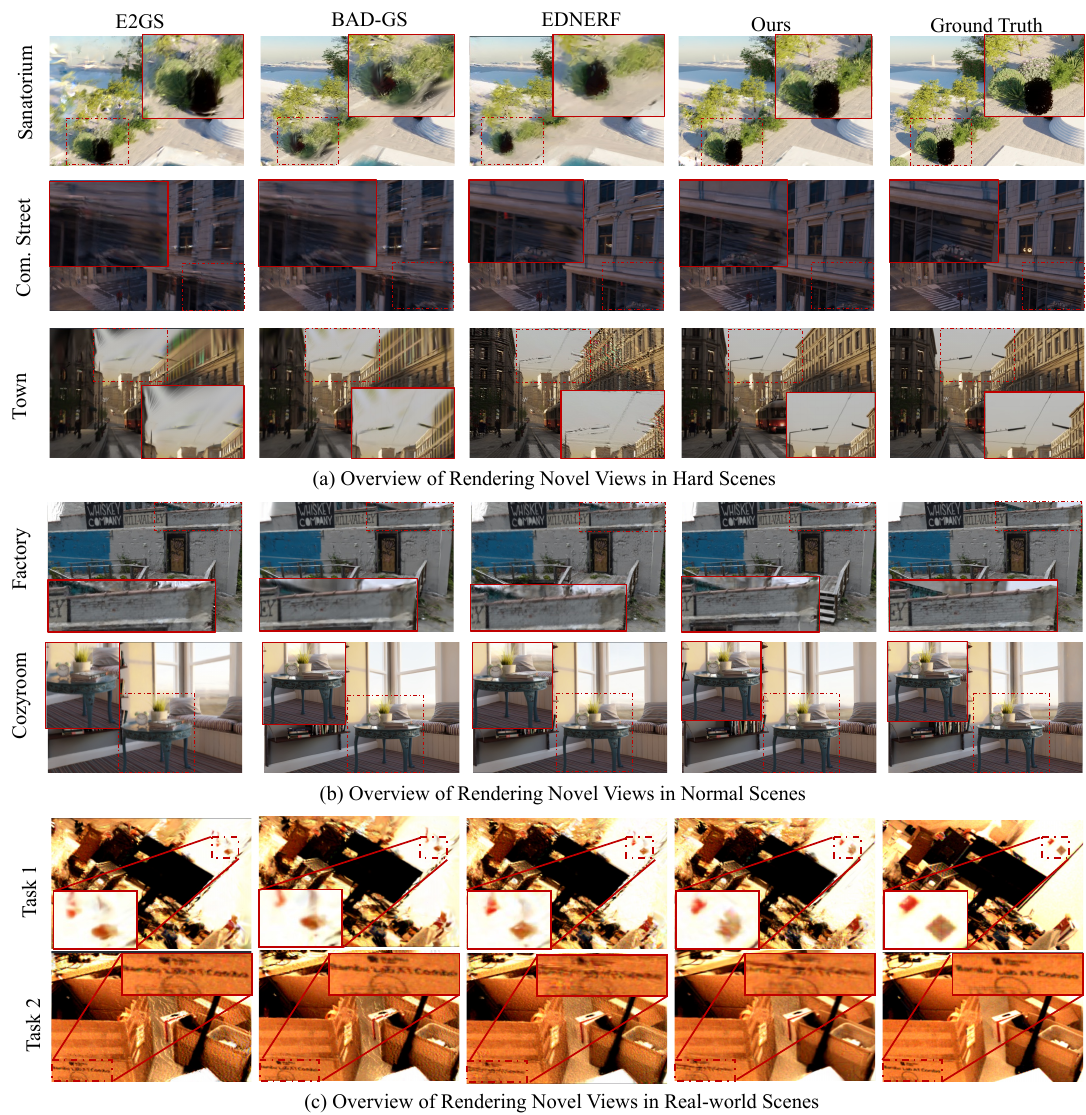}
    \caption{Qualitative comparison on the synthetic and real dataset. It shows that our method achieves better performance in rendering novel views compared to other SOTAs in deblur reconstruction. }
    \label{fig:mainpapersample}
    \vspace{-0.4cm}
\end{figure*}

\begin{figure*}[t]
    \centering
    \includegraphics[width=\linewidth]{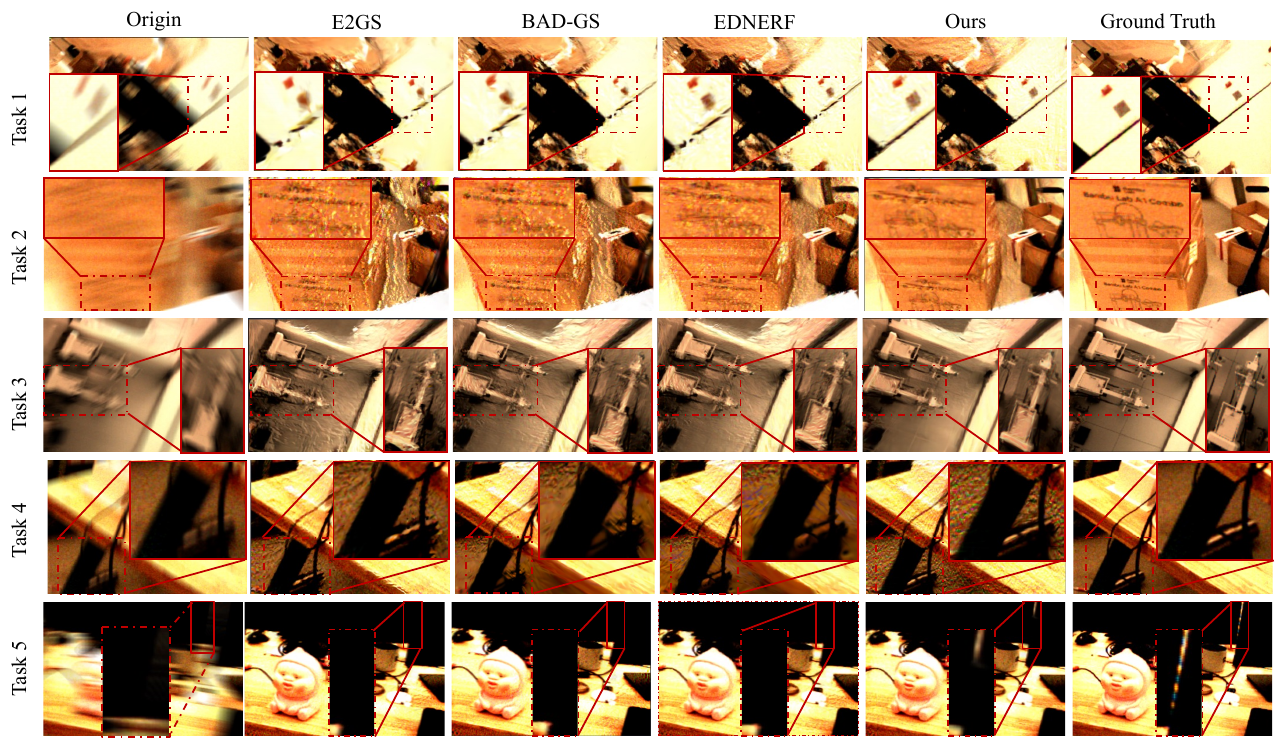}
    \caption{Qualitative comparison on the real-world scenes on deblur views rendering. It shows that our method achieves better performance in deblur views rendering compared to other SOTAs.}
    \label{fig:deblur-real}
\end{figure*}

\vspace{-0.2cm}
\subsection{Datasets}
Referring to \cite{Deblur-nerf,E2nerf}, we evaluate our method on both synthetic and real-world datasets. To comprehensively assess model performance, we introduce three types of benchmarks. In addition to normal dataset \cite{Deblur-nerf}, we consider more complicated scenes and propose the hard scenes crafted from Blender and real-world dataset. 
Without being limited to simple environmental conditions,
we also design the large-scale movement of camera motion, poses a great challenge to general deblurring reconstruction models. 

\subsubsection{Synthetic Dataset Details}
We construct a synthetic dataset with two difficulty levels, \emph{normal scenes} and \emph{hard scenes}, to systematically evaluate model performance under different scene complexities and motion conditions. All scenes are synthesized using Blender and follow the same rendering, motion blur, and event simulation protocol. \textbf{1. Normal Scenes.}  The normal-level dataset is adapted from BAD-NeRF~\cite{Bad-nerf} and consists of five scenes. These scenes exhibit relatively moderate object density, simpler layouts, and limited camera motion, serving as a standard benchmark for event-guided deblurring and reconstruction.
\textbf{2. Hard Scenes.}  To stress-test model robustness, we further construct seven challenging scenes using Blender, including both indoor and outdoor environments: Town, Sanatorium, Commercial Street, Boulevard, Archives, Restaurant, and Edifice (see Fig.~\ref{fig:scene}). Compared to the normal scenes, these hard scenes feature higher object density, more complex lighting conditions, richer texture details, and large-scale camera motions. The designed camera trajectories involve average displacements exceeding 10 meters, with dynamically shifting viewpoints across frames, posing significant challenges for both deblurring and 3D reconstruction.
For both normal and hard scenes, motion blur is synthesized by designing diverse camera trajectories. Specifically, we place multi-view cameras, randomly perturb their poses, and linearly interpolate between the original and perturbed positions to generate intermediate views. The final blurred images are obtained by rendering these interpolated views and blending them in the RGB space. Corresponding event streams are simulated using ESIM~\cite{V2E}. Following prior works~\cite{NeRF,Deblur-nerf}, we render 34 images per scene for both difficulty levels.

\subsubsection{Real-World Dataset with Multiple Levels of Motion Blur}
To evaluate performance in real-world scenarios, we manually recorded five scenes using the Color DAVIS346 event camera \cite{ESIM}, which features a resolution of $340 \times 260$ pixels and an exposure time of 100 milliseconds for RGB frames. To further facilitate the reality demand, we construct the this real-world blur datasets. We equip the DAVIS346 event camera on the UR-5E robotic arm as shown in Fig. \ref{fig:camera}. The goal of our dataset is to capture video frames and event data from the same scene with varying blur levels. To achieve this, we controlled the camera's motion using a robotic arm that performed nonlinear movements while adjusting the exposure time to induce different levels of blur. For accurate alignment of the videos with varying blur degrees, we used an IMU with a 1000 Hz sampling rate in combination with a registration algorithm to achieve millisecond-level temporal synchronization. We define 5 tasks and each task contains 5 kinds of blur in different exposure time (5e3, 5e4, 1.25e5, 2.5e5 and 5e5). We design a time interval alignment algorithm to match and correct all images and event streams. Finally, we utilize the blur degree of 5e3 as the ground truth sets and the scenes at other exposure times are used as a blur dataset. The images we captured are shown in Fig. \ref{10-RealBlurDVS}. We found that scenes with varying blur levels and limited image numbers (fewer than 30 images) often fail in initialization with COLMAP. Hence, we initialize these scenes by our method and simulate the sparse condition by random sampling 1k points on point clouds.


\begin{table}[t]
    \centering
    \caption{Analysis of DeblurSplat. ``Bala.'' denotes the Confidence Balanced Sampling. ``Prog.'' denotes Progressive Alignment.}
    \resizebox{0.9\linewidth}{!}{
    \begin{tabular}{cccccccc}
    \toprule
    \multirow{2}{*}{Bala.} & \multirow{2}{*}{Prog.} & \multicolumn{3}{c}{Hard} & \multicolumn{3}{c}{Normal} \\
                            \cmidrule{3-8}
                            & & PSNR$\uparrow$ & SSIM$\uparrow$ & LPIPS$\downarrow$ & PSNR$\uparrow$ & SSIM$\uparrow$ & LPIPS$\downarrow$ \\
    \midrule
    - & - &  22.29 & .6985 & .3037 & 24.98 & .7545 & .3123 \\
    - & \checkmark & 23.82 & .7621 & .1908 & 29.85 & .9273 & .0448 \\
    \checkmark & - & 23.32 & .7337 & .2547 & 26.19 & .8470 & .1130 \\
    \checkmark & \checkmark & \textbf{26.94} & \textbf{.8235} & \textbf{.1355} & \textbf{32.09} & \textbf{.9453} & \textbf{.0406} \\
    \bottomrule
    \end{tabular}}
    \label{tab:abl_frm}
\end{table}

\vspace{-0.2cm}
\subsection{Experiment Settings} 

\noindent \textbf{Baselines.} We compare our method with three types of baselines: 1) NeRF \cite{NeRF} and 3D-GS \cite{3DGS} directly trained on the blurry images, referring to as B-NeRF and B-3DGS. 2) Deblur rendering methods, including BAD-NeRF \cite{Bad-nerf}, BAD-GS \cite{Bad-gaussians}, $\text{E}^2$ GS \cite{E2gs} and EDNeRF \cite{Ev-GS}. 3) Image deblur methods, including UFP \cite{UFP} (single-image deblurring) and EFNet \cite{EFNet} (learnable event-based deblurring). We process input blurry images with them and train the vanilla 3D-GS with pre-deblurred images. 

\noindent \textbf{Evaluation Metrics.} We employ the Peak Signal-to-Noise Ratio (PSNR), Structural Similarity Index Measure (SSIM) \cite{ssim}, and VGG-based Learned Perceptual Image Patch Similarity (LPIPS) \cite{LPIPS} to evaluate the similarity between rendered views and ground truth views.

\subsection{Performance}

\noindent \textbf{Synthetic Data Experiments.} We evaluate our approach across a diverse set of synthetic scenes, including both normal and challenging (hard) scenarios. Quantitative results for novel view synthesis are presented in Table \ref{tab:all}, while detailed deblurring results of the input views can be found in the supplementary material. Our method demonstrates significant improvements across most metrics, particularly in challenging scenes, achieving a +2.74 dB increase in PSNR, a +5.78\% boost in SSIM, and a +6.24\% improvement in LPIPS. Specifically, both B-NeRF and B-3DGS generate blurry novel views as they are directly trained on blurred images. 
Compared to the BAD-Series methods \cite{Bad-nerf,Bad-gaussians}, our results highlight the advantages of leveraging event priors and dense stereo guidance. Additionally, while EDNeRF \cite{Ev-DeblurNeRF} attempts to decouple each blurry image into a single sharp prior, it fails to capture the continuous features and fine-grained variations in the event stream, ultimately leading to inferior performance.

\noindent \textbf{Real Data Experiments.} In this section, we analyze our model’s performance independently from the overall evaluation, focusing on different degrees of blur. As shown in Table \ref{tab:all}, our method outperforms state-of-the-art approaches, achieving a notable improvement of +3.36 dB in PSNR, +8.66\% in SSIM, and +6.24\% in LPIPS, demonstrating its practical applicability in real-world scenarios. We further investigate the impact of varying blur levels, as detailed in Table \ref{tab:blurlevel}. Our findings indicate that as the degree of blurring increases, the performance of all deblurring methods declines to varying extents. However, our approach consistently maintains superior performance, achieving a +2.15 dB gain in PSNR, highlighting its effective utilization of knowledge from DUSt3R and event streams. Additional results can be found in the supplementary.

\noindent \textbf{Qualitative Results.} Qualitative comparisons in Fig. \ref{fig:mainpapersample} and Fig. \ref{fig:deblur-real} demonstrate that \textbf{DeblurSplat} excels in recovering high-quality scene details under severe motion blur. While EDNeRF \cite{Ev-DeblurNeRF} tends to produce oversmoothed textures due to the limitations of implicit representations, existing 3DGS methods like E2GS \cite{E2gs} and BAD-GS \cite{Bad-gaussians} often collapse when traditional SfM or simplistic imaging models fail under extreme blur. In contrast, \textbf{DeblurSplat} provides a robust, SfM-independent initialization by integrating DUSt3R with event-camera priors. By further employing \textit{Progressive Alignment}, we leverage the high temporal resolution of events to iteratively refine latent sharp appearances and camera motion. This coarse-to-fine optimization enables our method to restore sharp edges and fine-grained textures in geometrically complex regions where other SOTA methods fail to maintain structural integrity.

\begin{table}[t]
    \centering
    \caption{Analysis of different sampling methods in initialization.}
    \resizebox{\linewidth}{!}{
    \begin{tabular}{ccccccc}
    \toprule
    \multirow{2}{*}{Method} & \multicolumn{3}{c|}{Hard} & \multicolumn{3}{c}{Normal} \\
                            \cmidrule{2-7}
                            & PSNR$\uparrow$ & SSIM$\uparrow$ & LPIPS$\downarrow$ & PSNR$\uparrow$ & SSIM$\uparrow$ & LPIPS$\downarrow$ \\
    \midrule
    Center  & 26.20 & .8078 & .1746 & 30.25 & .9296 & .0770 \\
    Random  & 26.63 & .8175 & .1386 & 31.84 & .9451 & .0413 \\
    Spatial  & 26.57 & .8166 & .1406 & 32.08 & \textbf{.9459} & \textbf{.0405} \\
    \textbf{Balanced}  & \textbf{26.94} & \textbf{.8235} & \textbf{.1355} & \textbf{32.09} & .9453 & .0406 \\
    \bottomrule
    \end{tabular}}
    \label{tab:samplingMet}
\end{table}

\begin{figure}
    \centering
    \includegraphics[width=\linewidth]{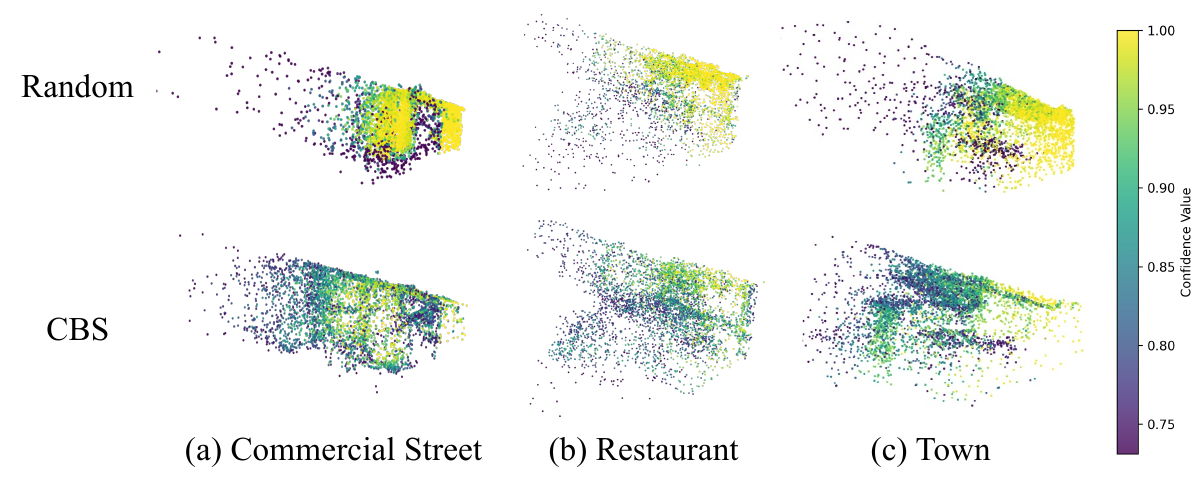}
    \caption{Illustration of point cloud comparison between Balanced Sampling (CBS) and random sampling.}
    \label{fig:sampling-shown}
\end{figure}

\subsection{Ablation Study}
In this section, we perform detailed study to investigate our designed modules in DeblurSplat by utilizing the NVS task. More ablation study can be referred to supplementary.

\noindent \textbf{Overview.} As shown in Table \ref{tab:all}, we first analyze the impact of each designed module. For our baseline, we adopt the standard 3DGS performance using only COLMAP initialization (B-3DGS). When incorporating only the "Bala" module, performance improves significantly over COLMAP initialization (+1.03 dB PSNR in hard scenes), benefiting from the fine-grained point clouds generated by a well-pretrained model. However, this approach still struggles with deblurring reconstruction, as it learns solely from blurred images without additional refinements. On the other hand, utilizing only the "Prog." module yields high-fidelity deblurred reconstructions, leveraging the rich semantic information from the event stream. Finally, by integrating both techniques, we achieve the best overall performance for NVS.

\begin{table}[t]
    \centering
    \caption{Analysis of weight $\lambda_e$ in optimization objectives.}
    \resizebox{\linewidth}{!}{
    \begin{tabular}{ccccccc}
    \toprule
    \multirow{2}{*}{$\lambda_e$} & \multicolumn{3}{c|}{Hard} & \multicolumn{3}{c}{Normal} \\
                            \cmidrule{2-7}
                            & PSNR$\uparrow$ & SSIM$\uparrow$ & LPIPS$\downarrow$ & PSNR$\uparrow$ & SSIM$\uparrow$ & LPIPS$\downarrow$ \\
    \midrule
    5e-3  & 26.94 & .8235 & .1355 & \textbf{32.09} & .9453 & \textbf{.0406} \\
    5e-2  & 23.14 & .7149 & \textbf{.1114} & 32.61 & \textbf{.9476} & .0471 \\
    5e-1  & 27.21 & .8357 & .1301 & 31.89 & .9420 & .0557 \\
    1.0  & \textbf{27.33} & \textbf{.8372} & .1298 & 31.24 & .9320 & .0670 \\
    \bottomrule
    \end{tabular}}
    \label{tab:weight}
\end{table}

\begin{table}[t]
    \centering
    \caption{Analysis of different sampling number in initialization.}
    \resizebox{\linewidth}{!}{
    \begin{tabular}{ccccccc}
    \toprule
    \multirow{2}{*}{Num.} & \multicolumn{3}{c}{Hard} & \multicolumn{3}{c}{Normal} \\
                            \cmidrule{2-7} 
                            & PSNR$\uparrow$ & SSIM$\uparrow$ & LPIPS$\downarrow$ & PSNR$\uparrow$ & SSIM$\uparrow$ & LPIPS$\downarrow$ \\
    \midrule
    5,000  & \textbf{26.94} & \textbf{.8235} & .1355 & \textbf{32.09} & \textbf{.9453} & \textbf{.0406} \\
    10,000  & 26.76 & .8198 & .1341 & 31.97 & .9459 & .0458 \\
    50,000 & 26.43 & .8070 & .1402 & 31.35 & .9336 & .0511 \\
    100,000 & 26.25 & .8012 & \textbf{.1064} & 30.73 & .9237 & .0553 \\
    \bottomrule
    \end{tabular}}
    \label{tab:samplingNum}
\end{table}

\begin{table}[t]
    \centering
    \caption{Analysis of interval $M$ numbers in initialization.}
    \resizebox{\linewidth}{!}{
    \begin{tabular}{ccccccc}
    \toprule
    \multirow{2}{*}{$M$} & \multicolumn{3}{c|}{Hard} & \multicolumn{3}{c}{Normal} \\
                            \cmidrule{2-7}
                            & PSNR$\uparrow$ & SSIM$\uparrow$ & LPIPS$\downarrow$ & PSNR$\uparrow$ & SSIM$\uparrow$ & LPIPS$\downarrow$ \\
    \midrule
    5  & 26.53 & .8141 & .1411 & 31.95 & .9442 & .0424 \\
    10  & 26.81 & .8188 & .1383 & 31.92 & .9442 & \textbf{.0357} \\
    20  & 26.86 & .8222 & \textbf{.1328} & 31.11 & \textbf{.9463} & .0404 \\
    40  & \textbf{26.94} & \textbf{.8235} & .1355 & \textbf{32.09} & .9453 & .0406 \\
    80  & 26.73 & .8188 & .1399 & 32.00 & .9439 & .0419 \\
    \bottomrule
    \end{tabular}}
    \label{tab:interval}
\end{table}

\begin{table}[t]
    \centering
    \caption{Analysis of the proposed modules for pose estimation. The results are in the absolute trajectory error metric (ATE). Note that COLMAP initialization is based on sharp images, while our method utilize blur images.}
    \resizebox{\linewidth}{!}{
    \begin{tabular}{lcccc}
        \toprule
        \multirow{2}{*}{Methods} & \multicolumn{2}{c}{Before Training} & \multicolumn{2}{c}{After Training} \\
        \cmidrule{2-3}  \cmidrule(lr){4-5}
        & COLMAP & Ours & Ours w/o event & Ours w/ event \\
        \midrule
        Average     & 1.7590  & \textbf{0.2742}  & 0.0560  & \textbf{0.0471} \\
        \midrule
        Archives    & 1.1187  & \textbf{0.1459}  & 0.1127  & \textbf{0.0964} \\
        Edifice     & 2.6226  & \textbf{0.0181}  & 0.0292  & \textbf{0.0171} \\
        Com. Street & 1.1329  &\textbf{ 0.0949}  & 0.0632  & \textbf{0.0582} \\
        Restaurant  & 1.1627  & \textbf{0.0536}  & 0.0214  & \textbf{0.0111} \\
        Sanatorium  & 2.7585  & \textbf{0.3272}  & 0.1312  & \textbf{0.1192} \\
        Town        & 1.9840  & \textbf{0.7073}  & 0.0203  & \textbf{0.0177} \\
        Boulevard   & 1.6330  & \textbf{0.5720}  & 0.0183  & \textbf{0.0079} \\     
        \bottomrule
    \end{tabular}}
    \label{tab:ATE}
\end{table}

\begin{table}[t]
    \centering
    \caption{Ablation study of Grayscale Guidance in Hard Scenes.}
    \begin{tabular}{lccc}
    \toprule
    Guidance & PSNR$\uparrow$ & SSIM$\uparrow$ & LPIPS$\downarrow$ \\
    \midrule
    Chromatic & 26.71 & .8192 & .1381 \\
    \textbf{Grayscale  (Ours)} & \textbf{26.94} & \textbf{.8235} & \textbf{.1355} \\
    \bottomrule
    \end{tabular}
    \label{tab:gray_guidance}
\end{table}

\noindent \textbf{Sampling Techniques.} We investigate the impact of different sampling methods for generating point clouds, as shown in Table \ref{tab:samplingMet}. The "Spatial" method employs Farthest Point Sampling (FPS) within adaptive voxel grids \cite{Instantsplat}. The "Random" method directly samples points from the point cloud without any constraints. The "Center" method selects a high-confidence point cloud center and randomly samples points within its average geometric distance. Our findings indicate that the "Center" method leads to performance degradation, as it filters out distant points that play a crucial role in enhancing generalization ability. Meanwhile, the "Random" method, despite the possibility of sampling less relevant points, maintains high NVS quality. This supports our hypothesis in Section \ref{subsec:pa}—that even lower-confidence points can carry meaningful semantics and contribute to performance. Although the "Spatial" method performs well in simple scenes, it struggles in more challenging scenarios. By relying on geometric distance, it increases the likelihood of sampling less informative points, ultimately limiting its robustness in complex environments. We also investigate the different point cloud quality between Balanced Sampling (CBS) and random sampling as shown in Fig. \ref{fig:sampling-shown}. The heatmap clearly illustrates that under random sampling, the point cloud is highly concentrated in specific "yellow" (high-confidence) zones, leaving peripheral areas sparse and increasing computational load without adding new spatial information. This proves that CBS strikes a critical balance: it prevents stochastic redundancy while guaranteeing the robust geometric backbone that random sampling leaves to chance.

\begin{figure}[t]
    \centering
    \includegraphics[width=0.9\linewidth]{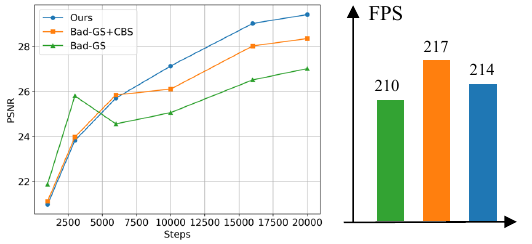}
     \vspace{-0.2cm}
    \caption{Analysis of efficiency of Confidence Balanced Sampling (CBS) in Hard Scenes. Here, we utilize the PSNR in deblur view rendering as the evaluation metric.}
    \label{fig:efficiency}
    \vspace{-0.2cm}
\end{figure}

\noindent \textbf{Initialization and Rendering Efficiency.} 
We first evaluate the efficiency of the proposed Confidence Balanced Sampling (CBS) module from two complementary perspectives: training convergence and rendering speed. As shown in Fig.~\ref{fig:efficiency}, integrating CBS into BAD-GS~\cite{Bad-gaussians} significantly accelerates training convergence. Specifically, our method reaches a PSNR comparable to BAD-GS trained for 20,000 iterations using only 12,500 training steps, demonstrating a clear improvement in initialization and optimization efficiency. Importantly, this gain in training efficiency does not come at the cost of inference performance. All evaluated configurations maintain real-time rendering speeds exceeding 210 FPS, which is on par with the BAD-GS baseline. This indicates that leveraging dense geometric priors from DUSt3R, together with CBS-based initialization, improves convergence behavior without introducing additional computational overhead during the rasterization stage.

\noindent \textbf{Optimization Objectives.} We explore different $\lambda_e$ settings from Eq. \ref{eq:opm} to assess model performance. As shown in Table \ref{tab:weight}, we conclude that event-based guidance and latent blur simulation are crucial for reconstruction performance in challenging scenarios when $\lambda_e \geq$ 5e-1. On the other hand, setting $\lambda_e$ = 5e-2 leads to performance collapse, likely due to confusion between the event-based prior and the original blurred images, causing the model to fall into sub-optimal solutions.

\noindent \textbf{Initialization Efficiency.} We analyze the effect of different initialization interval values ($M$), as presented in Table \ref{tab:interval}. Our findings indicate that setting $5 \leq M \leq 20$ hinders fine-grained semantic extraction, while increasing $M$ to 80 overly prioritizes low-confidence points, many of which lack meaningful information. Finally, we examine the impact of different sampling quantities on overall optimization. As shown in Table \ref{tab:samplingNum}, we observe that increasing the sampling quantity during initialization makes global optimization more challenging, aligning with findings from previous works \cite{Pointnet++,3DGS}. To further improve the reconstruction performance and robust deblurring, we set $M$ to 40.



\noindent \textbf{Pose Estimation.} As shown in Table \ref{tab:ATE}, we employ the Absolute Trajectory Error (ATE) metric to assess the impact of different modules on pose estimation. Compared to our proposed approach, COLMAP initialization, which relies on sharp images, produces significantly higher errors. This is attributed to its limited robustness when handling blurred images, where semantic confusion degrades pose estimation accuracy. In contrast, our method leverages the rich priors encoded in a dense stereo module pretrained on large-scale sharp datasets, enabling more accurate camera pose predictions. This proves that our pose optimization module
not only recovers from disadvantaged initializations but also reaches a level of precision that exceeds
traditional pipelines given ideal inputs. Furthermore, the results in Table \ref{tab:ATE} demonstrate that the Progressive Alignment module effectively refines pose estimation by implicitly guiding optimization in challenging scenarios, ultimately contributing to improved deblurring reconstruction.

\noindent \textbf{Grayscale Guidance.} We analyze the impact about the grayscale guidance of Eq. \ref{eq:fine}. Motion blur is a spatially varying yet channel-consistent degradation, as the underlying camera trajectory is identical across RGB channels. Since events are triggered by changes in log-intensity, grayscale events capture the essential structural and motion cues required for deblurring, while avoiding channel-wise noise amplification introduced by color-filter responses. As shown in Table \ref{tab:gray_guidance}, grayscale guidance consistently outperforms chromatic guidance across all metrics, yielding higher PSNR and SSIM as well as lower LPIPS, particularly in hard scenes. This confirms that luminance-based event supervision provides a more robust and effective guidance signal.

\section{Limitation and Future Work}

\noindent \textbf{Confidence Evaluation.} We rely on confidence estimation to filter point clouds generated by DUSt3R. However, its accuracy under ambiguous conditions remains unclear, especially given that DUSt3R is trained on high-quality, sharp images. A more rigorous quantitative standard for assessing point cloud quality is needed to ensure reliable initialization of gaussian primitives.

\noindent \textbf{Large-Scale Test.} The core concept of our approach leverages the rich knowledge from DUSt3R and event streams to tackle motion-blur reconstruction. Naturally, this raises the question of whether the same mechanism can be extended to more complex reconstruction scenarios, such as rolling-shutter effects.

\noindent \textbf{Resolution Limitations.} In our real-world experiments, we employ a DAVIS346 sensor with a spatial resolution of 346$\times$260, which is lower than that of modern high-definition RGB cameras. 
Our work primarily focuses on the fidelity of motion deblurring and novel view synthesis (NVS) under complex motion trajectories, rather than addressing the orthogonal challenge of cross-resolution super-resolution. 
While the achievable reconstruction quality is bounded by the native resolution of the event sensor, leveraging low-resolution event streams to guide high-resolution reconstruction remains a promising and non-trivial direction. 
We leave the joint modeling of event-guided deblurring and resolution enhancement for future work.

\noindent \textbf{EDI Assumptions.} 
Our Progressive Alignment module relies on the Event-based Double Integral (EDI) model \cite{EDI} to decouple motion blur into latent sharp frames. This formulation implicitly assumes (1) a constant contrast threshold $\Theta$ across pixels, (2) locally static scene content during exposure (i.e., blur is primarily caused by camera motion), and (3) relatively clean event measurements. In real-world scenarios, however, these assumptions may be partially violated due to sensor noise, spatially varying contrast thresholds, or non-uniform illumination changes. In particular, event cameras exhibit pixel-wise threshold variability and background activity noise, which may introduce bias in the reconstructed latent intensities. Moreover, sudden lighting flicker or dynamic objects may not strictly satisfy the constant-velocity or static-scene assumptions embedded in Equation \ref{eq:linearpose}. Although our framework partially alleviates these issues by jointly optimizing Gaussian primitives and latent poses under multi-view photometric supervision, the physical simplifications of EDI may still limit reconstruction fidelity under extreme lighting transitions or highly dynamic scenes. In future work, we plan to explore adaptive threshold modeling, noise-aware event filtering, and dynamic-scene extensions to further improve robustness beyond the standard EDI formulation.

\section{Conclusion} In this paper, we introduce DeblurSplat, the first traditional SfM-free deblurring framework for 3D-GS that integrates event-camera priors with DUSt3R’s robust initialization. By leveraging Confidence Balanced Sampling, we ensure reliable point cloud initialization, while Progressive Alignment effectively refines motion-blurry pixels using event-stream semantics. Our method outperforms previous motion-deblur reconstruction approaches across diverse synthetic and real-world datasets, demonstrating superior robustness under varying blur levels. These results highlight the effectiveness of SfM-free deblurring and event-based priors in enhancing 3D scene reconstruction.

{   
    \small
    \bibliographystyle{ieeenat_fullname}
    \bibliography{main}
}

\clearpage

\end{document}